%% file: paper.tex
\documentclass[runningheads]{llncs}
\usepackage[T1]{fontenc}
\usepackage{graphicx}
\usepackage[utf8]{inputenc} 
\usepackage[T1]{fontenc}    
\usepackage{hyperref}       
\usepackage{url}            
\usepackage{booktabs}       
\usepackage{amsfonts}       
\usepackage{nicefrac}       
\usepackage{microtype}      
\usepackage{xcolor}  
\usepackage{tabularx}

\begin{document}
\title{Adesua: Development and Feasibility Study of an AI WhatsApp Bot for Science Learning in West Africa}
\titlerunning{Adesua}
%

\author{George Boateng\inst{1,3}  \and
Evans Atompoya \inst{3}  \and
Philemon Badu \inst{3} \and
Samuel John \inst{3}  \and
Samuel Ansah \inst{3} \and
Patrick Agyeman-Budu \inst{3} \and
Victor Kumbol \inst{2,3}}


\authorrunning{G. Boateng et al.}
\institute{ETH Zurich, Switzerland\\
\and
Charité - Universitätsmedizin Berlin, Germany\\
\and
Kwame AI Inc., U.S.}


\maketitle 


\begin{abstract}
Sub-Saharan Africa faces persistently high student–teacher ratios and shortages of qualified teachers, limiting students’ access to personalized learning support and formative assessment. To address this challenge, we present Adesua, a WhatsApp-based AI Teaching Assistant for science education that extends the Kwame for Science platform. Adesua leverages WhatsApp’s widespread adoption in Africa to provide accessible, curriculum-aligned learning support for Junior High School (JHS) and Senior High School (SHS) students across West Africa. The system integrates curated textbooks and 33 years of national examination questions with generative AI to enable conversational question answering and automated assessment with feedback via a WhatsApp bot. Students can ask science questions, take timed or untimed multiple-choice tests by topic or exam year, and receive instant grading and detailed explanations of correct and incorrect responses. A 6-month feasibility deployment in 2025 had 56 active users in Ghana, including students and parents. Quantitative evaluation showed a high perceived usefulness, with a helpfulness score of 93.75\% for AI-generated answers, albeit with a small number of ratings (n=16). These preliminary results provide a basis for more extensive future evaluation of a WhatsApp-based AI assistant to assess its potential to offer scalable, low-cost personalized learning support and formative assessment in resource-constrained educational contexts.

\end{abstract}

\input{content.tex}


\bibliographystyle{splncs04}
\bibliography{refs}

\end{document}

%% file: content.tex
\section{Introduction}
In Sub-Saharan Africa, the education sector faces significant challenges due to a shortage of qualified teachers and high student-teacher ratios. As of 2022, only 69\% of primary school teachers in the region met the minimum required qualifications, a decline from 75\% in 2010 \cite{UNESCO2024}. This shortage is further exacerbated by the need to recruit an additional 15 million teachers by 2030 to achieve universal primary and secondary education \cite{UNESCO2021}. Consequently, students often lack personalized learning support, such as timely question-answering and formative assessments, which are crucial for monitoring and enhancing their educational progress.

To address this challenge, Boateng et al. developed Kwame for Science \cite{boateng2023b}, an AI-powered web app that enables high school students across West Africa to get answers to questions in the science curriculum and view past national exam questions.  Kwame for Science, however, has some limitations. Firstly, it only returns passages from curated textbooks as answers rather than direct answers to questions. Consequently, learners have to infer answers from the passages. Secondly, it lacks a test-taking feature, so learners cannot assess their understanding of various concepts. Finally, Kwame for Science, as a web app, is not very accessible for the target population of West Africans, for whom Internet data is expensive \cite{dw2020}.

Recent advances in generative artificial intelligence (AI), particularly large language models (LLMs), enable new forms of personalized educational support that can help address some of the limitations of Kwame for Science. This paper extends Kwame for Science and makes the following contributions: (1) the design and implementation of a retrieval-augmented generation (RAG) question-answering system that produces direct, curriculum-aligned science explanations grounded in curated West African exam and textbook content; (2) an interactive assessment framework delivered via WhatsApp that supports both premade and custom quizzes with automated grading and detailed feedback; (3) the integration of these capabilities into \textbf{Adesua}, a low-bandwidth, messaging-based educational platform tailored for Junior High School (JHS) and Senior High School (SHS) science education in West Africa; and (4) empirical insights from a six-month feasibility deployment with 56 users in Ghana that illustrate usage patterns and perceived usefulness of the system. Together, these contributions demonstrate how generative AI can be operationalized to provide accessible, personalized science learning support in resource-constrained contexts.

\section{Background and Related Work}
\subsection{Kwame for Science}
Kwame for Science is an AI-powered web app that enables students across West Africa to access two features: (1) question answering and (2) view past questions \cite{boateng2023b}. The question answering component displays 3 passages from a curated knowledge base of textbooks as answers to students’ questions, along with the top 5 past exam questions (and expert answers) that are related to their questions. When a student asks a question, the system extracts an embedding from the text using a Sentence-BERT (SBERT)  model \cite{reimers2019}, computes cosine similarity between the embedding and the pre-stored embeddings of passages from textbooks stored in ElasticSearch on Google Cloud Platform, and then returns the top passage as an answer based on the cosine similarity scores. Students can rate the helpfulness of answers and related questions. Kwame for Science is enabled by a curated knowledge base of content from textbooks and past national exams over the past 28 years for Integrated Science at the SHS level, with answers from certified teachers. The View Past Questions feature allows students to search and view past national exam questions and answers with filters for examination year, specific exam, question type, and automatically categorized topics generated by a custom topic detection model. The topic detection model is a machine learning model that was trained using embeddings computed with SBERT and a Support Vector Machine to automatically categorize questions into topics in the syllabus for all 28 years of exams. During an 8-month deployment of Kwame for Science from June 2022 to February 2023, there were 750 users who asked 1.5K questions, with Kwame’s helpfulness score being 87.2\% \cite{boateng2023b}. 

\subsection{Chatbots for Science Education}
Several studies have applied and evaluated chatbots as tutoring support for high school science students \cite{lin2023,deveci2021,lee2023,chang2023}. Some of these chatbots have been deployed and evaluated in Africa. FoondaMate, is an AI chatbot accessible via WhatsApp and Facebook Messenger, which provides students with past national exams and supports them with their homework \cite{foondamate}. Similarly, Rori, is an AI-powered virtual math tutor accessible via WhatsApp, which delivers micro-lessons and poses practice questions, enabling learners to progress at their own pace \cite{rori,henkel2024}. 

Rori operates on a pedagogical model rather than the question-bank approach adopted by Foondamate. By delivering structured microlessons using the Teaching at the Right Level methodology, Rori is designed to teach math concepts from scratch to students in Grades 3-9. In addition, Rori is integrated into school systems which enables institutional scaling with teacher support \cite{henkel2024}. Foondamate has a wider coverage of over 30 subjects at Grades 8-12 including accounting and languages aligned with the South African CAPS curriculum \cite{foondamate}.  There is no published quantitative data on accuracy or helpfulness for Foondamate but several user testimonials have been documented. Rori demonstrated an improvement in test scores with an effect size of 0.36 SD when combined with normal math lessons in an evaluation involving 1,000 high school students who used Rori twice weekly over 8 months \cite{henkel2024}.

These examples show that AI chatbots provide students with instant and individualized feedback \cite{chen2024,taani2025},  increase students’ interest and motivation by making learning more engaging \cite{lee2023,deveci2021}, frees up teachers’ time spent in grading assessments \cite{kurniawan2024}, resulting in an overall improvement in learning outcomes \cite{lee2023}. While there remain challenges such as inaccuracy of AI responses,  insufficient digital literacy of students and teachers, and lack of access to devices, which limit the effective use of these chatbots \cite{fayzullina2025}, these studies highlight the potential of AI chatbots to provide scalable, personalized learning support, especially in limited resource settings. 

In contrast to prior educational chatbots such as FoondaMate and Rori, Adesua uniquely combines retrieval-augmented generative question answering with curriculum-aligned, exam-based assessments and detailed feedback within a single system for Sceience learning. By grounding generative responses in curated expert-verified answers to 33-years of local exams as well as textbook content aligned to the West African Science education curricula for JHS and SHS (Grades 7-12), Adesua stands out as an integrated, context-aware learning and assessment platform tailored specifically to West African Science education.

\section{System Overview}
\begin{figure}[t]
  \centering
  \includegraphics[width=\linewidth]{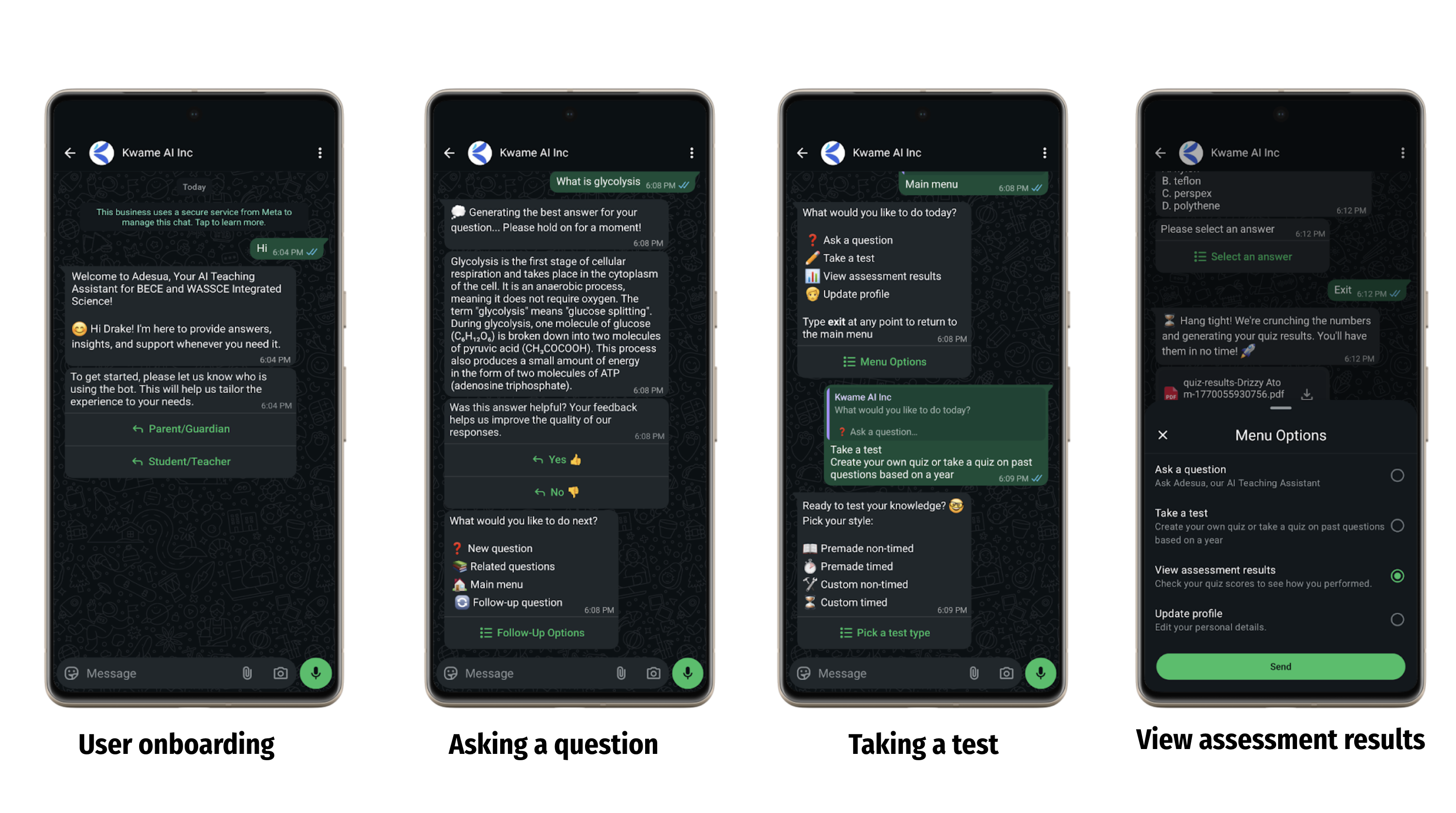}
  \caption{Screenshots of Adesua}
  \label{fig:adesua_pilot}
\end{figure}

Adesua \footnote{See \href{https://adesua.kwame.ai/}{Adesua Website} and   \href{https://res.cloudinary.com/dsul5wugf/video/upload/Adesua_Bot_atbfhe.mp4}{Demo Video}} (Figure \ref{fig:adesua_pilot}) is an AI-powered WhatsApp bot for personalized question-answering and assessment for science education in West Africa.  It provides educational support to students through WhatsApp, leveraging the ubiquity of mobile messaging and low Internet data consumption to deliver personalized learning experiences. The system is designed to support students at both JHS and SHS levels across West Africa, providing access to educational content, assessment tools, and personalized feedback mechanisms. The content that powers Adesua consists of the past 33 years of national exam questions at the JHS level (1990 to 2023) that we curated and annotated to add to our previously curated 28 years of SHS exam questions. We then used GPT-4 API to generate answers to all of the questions and had an expert verify the answers.  Every GPT-4 answer in the dataset was reviewed by the expert. During this review, the expert checked each AI-generated answer against the corresponding question and recorded their judgment in a dedicated "Expert Answer" column in the dataset spreadsheet. Where GPT-4's answer was incorrect, the expert provided the correct response in that column. This means the human-in-the-loop validation covered 100\% of the dataset, and the reviewer's answer column serves as the ground-truth reference used for all downstream evaluation. The Adesua system provides four primary functions through its conversational interface that enable them to onboard and navigate, ask questions, complete assessments, and review their performance. 

\subsection{Onboarding and Navigation}
All users must complete an onboarding process before accessing these features. The system determines whether the user is a parent, guardian, student, or teacher, collects appropriate consent in accordance with user age and relationship to the student, ensures agreement to the terms of service, and guides the student through providing necessary profile information, including school affiliation and education level, before granting access to the main functionality. The system incorporates robust error handling and user guidance mechanisms to support effective interaction. The exit command provides a global navigation mechanism accessible from most conversation states, except for onboarding flows and active quiz sessions, where completion or explicit cancellation is required. When users submit input that cannot be interpreted within the context of their current conversation step, the system provides clarifying guidance. Rather than proceeding with potentially incorrect assumptions, the system reminds users of the available options for their current step and re-presents the most recent prompt. This approach helps users understand the expected input format and reduces frustration from misunderstood interactions. Throughout all interactions, the system maintains contextual awareness of the conversation state and responds appropriately to the user's current position in the interaction flow. This state-based design ensures that users receive relevant prompts and that their inputs are interpreted correctly according to their current task.

\subsection{Question and Answer System}
The question and answer feature enables students to submit educational queries in natural language and receive instand AI-generated responses grounded in curated, local, curriculum-aligned Science content tailored to their educational level. It uses the Basic Education Certificate Examination (BECE) curriculum for JHS students and the West African Secondary School Certificate Examination (WASSCE) curriculum used by Ghana, Nigeria, Gambia, Sierra Leone, Liberia, and The Gambia for SHS students.

When a student selects the "Ask Question" option from the main menu, the system preprocesses the query by converting it to lowercase and stripping punctuation. Queries shorter than five characters are rejected to prevent low-intent or noisy retrievals. The system employs a RAG pipeline designed to minimize LLM hallucinations by anchoring all responses in verified source material before invoking a generative model. The pipeline proceeds in three stages: semantic retrieval, context assembly, and answer generation.

\textbf{Semantic Retrieval:} The system uses the same retrieval pipeline as Kwame for Science \cite{boateng2023b}. The preprocessed query is transformed into a high-dimensional vector representation using the all-mpnet-base-v2 sentence embedding model. The system then executes two concurrent ElasticSearch queries using a script\_score configuration with cosine similarity as the relevance metric. The first query searches a general science content index that contains passages from open-source Science textbooks; the second searches an exam-specific index corresponding to the student's curriculum level — BECE for JHS students or WASSCE for SHS students. A relevance threshold of 0.6 is applied to both indices, and only results meeting this threshold are retained. The top-5 retrieved passages and top-5 related past exam questions are returned for downstream use. The pipeline exposes granular latency telemetry at each stage: embedding time, per-index retrieval time, data restructuring overhead, and end-to-end response latency. In our deployment, median end-to-end latency was approximately 1.8 ms.

\textbf{Context Assembly:} The retrieved passages and related exam questions are assembled into a structured prompt context. Up to eight context slots are populated: three top-ranked textbook passage excerpts and five related past exam question-answer pairs. Figures, code blocks, and tables embedded in retrieved passages are detected via structural markers and resolved to their respective Google Cloud Storage asset URLs before being passed downstream.

\textbf{Answer Generation:}. The assembled context, together with the student's original query, is passed to GPT-4 via Azure OpenAI with a curriculum-specific system prompt. The system prompt instructs the model to answer accurately and concisely for Ghanaian JHS or SHS students, use WhatsApp-compatible formatting (single asterisks for emphasis, Unicode superscripts and subscripts for chemical equations), avoid LaTeX syntax, and cap responses at 4,096 characters. The model is configured with a temperature of 1.0, top-p of 1.0, and a maximum of 1,024 output tokens. The prompt explicitly directs the model to rely on the provided contexts and, where contexts are incomplete, to default to a level-appropriate explanation rather than fabricate information.

This pipeline represents a significant architectural advancement over the original Kwame for Science system, which returned raw passage excerpts directly to the user without generative synthesis. In the prior system, learners were required to infer answers from retrieved text. Adesua's RAG architecture instead produces a single, coherent, curriculum-calibrated response while preserving factual grounding in verified source material — addressing the core limitation of the predecessor system.

Upon receiving the generated answer, the student may ask a new question, return to the main menu, request related questions with answers, or submit a follow-up question. For follow-up questions, the system maintains prior conversation turns to enable coherent multi-turn educational dialogues. A feedback mechanism allows students to rate the helpfulness of each answer; these ratings are logged for quality assurance and future model improvement.

\subsection{Assessment System}
Students can take timed or non-timed tests (multiple choice questions) on past exam years, or specific topics by selecting a topic and receiving questions. The answers provided are automatically graded, and students are given a detailed report and feedback on correct and incorrect answers to improve their understanding. The assessment system offers two distinct quiz modalities: premade quizzes and custom quizzes, each available in timed and untimed variants. Premade quizzes are based on previous examination papers organized by year and subject, providing students with authentic assessment materials aligned with national curriculum standards. Custom quizzes are dynamically generated from specific topics identified by the student, offering targeted practice in particular subject areas. When a student selects the premade quiz option, the system presents a list of available assessments filtered by the student's education level. The student can describe their desired quiz in natural language, such as 'integrated science WASSCE 2023,' and the system employs natural language processing to identify the matching quiz and confirm the selection. For custom quizzes, the system requests the student to specify a topic. It then matches this input to pre-stored curriculum topics and retrieves an appropriate set of objective questions. In both cases, students selecting the timed variant receive a predetermined time allocation based on the number of questions in the assessment. Once a quiz is selected, the system requests explicit confirmation from the student to begin. During the assessment, questions are presented sequentially, with each question potentially including both textual content and supporting images. Students respond by selecting from multiple-choice options labeled A through F. For timed assessments, a countdown mechanism runs for the duration of the quiz, and the assessment automatically concludes when time expires, even if not all questions have been answered. Students can exit an assessment before completion using either a provided button or by typing 'exit.' This returns them to the quiz type selection without submitting their responses. Upon completion of all questions or expiration of the timer, the system grades the assessment, generates a comprehensive PDF report containing the questions, student responses, correct answers, and explanatory content, and delivers this document to the student via WhatsApp. The assessment result is also stored in the student's profile for later access through the results viewing feature. Following the delivery of results, the student is returned to the main menu.

\subsection{Assessment Results Repository}
The assessment results feature provides students with access to their complete testing history. When selected from the main menu, the system retrieves all completed assessments for the student. If results exist, the system generates and delivers a PDF document that lists each assessment with a hyperlink to its detailed result PDF. This enables students to review their performance across multiple assessments and track their progress over time. If no assessment results exist in the student's profile, the system informs the student of this status and returns them to the main menu. This design ensures that students receive clear feedback regardless of their assessment history.

\begin{figure}[t]
  \centering
  \includegraphics[width=\linewidth]{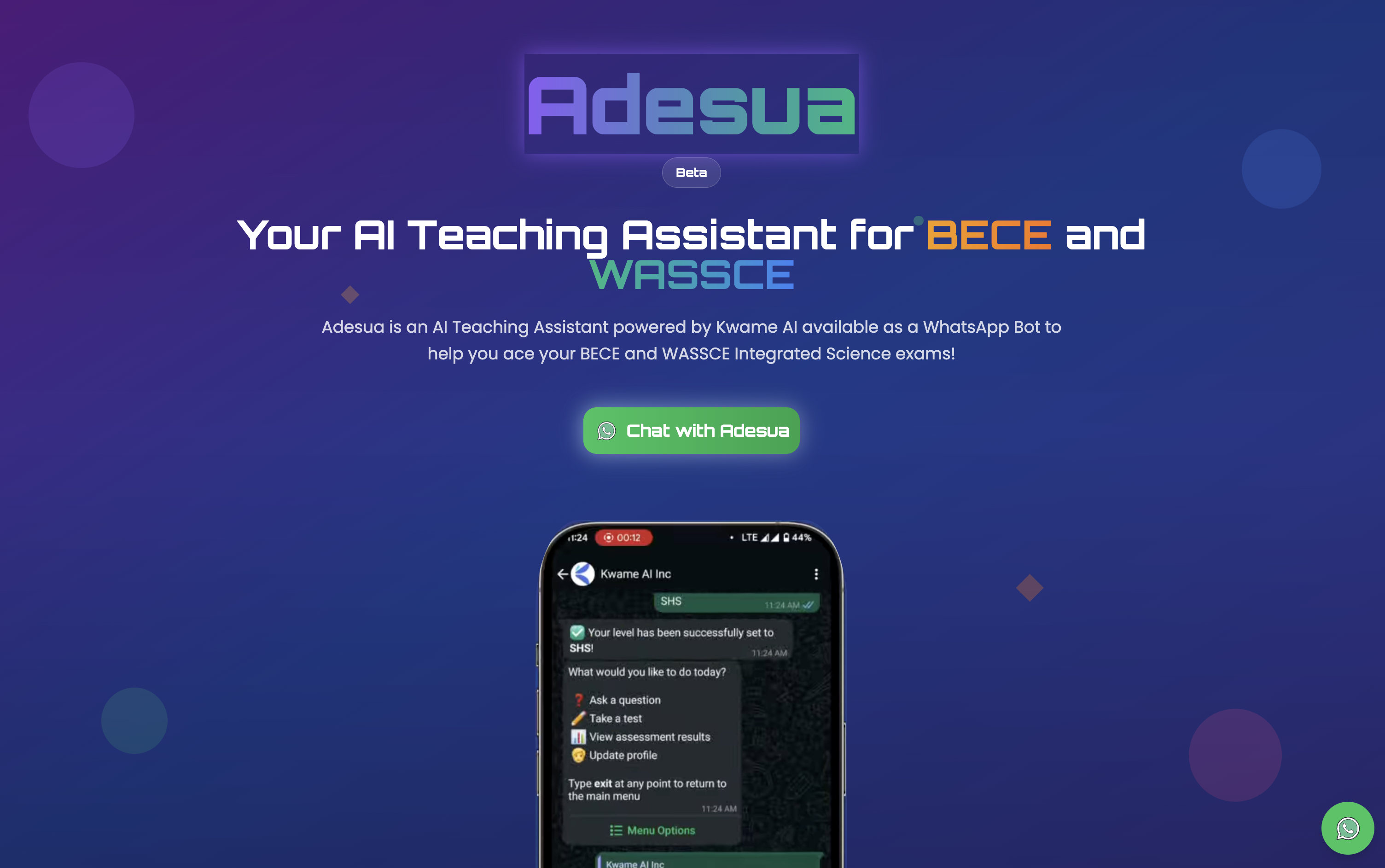}
  \caption{Adesua Landing Page}
  \label{fig:landing_page}
\end{figure}

\section{Feasibility Study and Evaluation}
We launched Adesua in 2025 and ran a study over 6 months. We publicized the tool on social media, sharing a catchy landing (Figure \ref{fig:landing_page}). We also collaborated with a teacher organization in Ghana to publicize the tool to their teachers, who would then share the information with their students. Our evaluation was preliminary. We performed quantitative evaluation. We also got feedback from some users and used data from our internal users' logs to fix bugs and improve the tool. In total, we had 107 signups — the number of people who sent an initial message to the bot and whose names and numbers were stored. Out of this number, 56 users moved to the next stage of indicating their status as parents or students (18 were parents, and 38 were students). Only 46 users provided an educational level, out of which 30.4\% (14) were at the JHS level, and 69.6\% (32) were at the SHS level. 

A total of 44 assessment attempts were recorded across 19 unique users. Of these, 20 (45.5\%) were completed, 22 were abandoned before completion, and 2 were not started. Thirteen (13) unique users completed at least one assessment. Completion rates also varied by quiz type: custom topic quizzes, which are shorter and contain a maximum of 20 questions, achieved higher completion rates than full past-paper exams, which contain 40–60 questions. Among incomplete attempts, the majority were abandoned very early in the session, with most users answering fewer than 10\% of questions before dropping off.

A total of 16 users made 48 queries in total and could respond yes or no to the question "Was this answer helpful?" in response to the answers they received from our AI. We computed the helpfulness score of our Ask Kwame feature, which was 93.75\% (n=16), indicating that the answers were very helpful, albeit with a small sample. These results are preliminary; hence, a larger, more extensive study is needed before any conclusions can be made.

\section{Limitations and Future Work}
Our collaboration was not seamless and had challenges, resulting in a lack of a structured group of students who used Adesua, resulting in limited engagement on the platform. We hypothesize that a good number of those who signed up were merely curious about the platform as opposed to wanting to use it for learning, which resulted in low engagement. One key lesson is to establish formal partnerships with proactive organizations that work with schools for deployments like these. Also, we didn't evaluate the impact of the tool on learning outcomes. In the future, we will evaluate the tool's utility and effect on learning outcomes, such as local and national exams, using a randomized control trial. Additional features we have in mind to improve the tool are as follows: (1) Allow students to upload images of questions and get answers to them, (2) Add an assessment for open-ended questions, (3) Integrate speech-to-text, local translation and text-to-speech systems to allow students to send voice notes and get voice responses to their questions in a Ghanaian accent and local languages, and (4) Implement an AI tutoring component of the tool to tutor students on specific topics in the Ghanaian Science syllabus.

\section{Conclusion}
In this work, we built Adesua, an AI-powered WhatsApp bot that extends Kwame for Science by providing direct, curriculum-aligned question answering, interactive assessments with automated feedback, and improved accessibility through a widely used messaging platform. By integrating RAG with curated local science content, Adesua demonstrates how generative AI can be adapted to support personalized learning in resource-constrained educational contexts. Results from a 6-month feasibility study with 56 users in Ghana suggest that students found the system helpful, albeit, based on a small number of votes (n=16). While these findings are preliminary and based on a small sample, they provide a basis for future extensive evaluation of WhatsApp-based AI tools with a structured cohort of students to assess how the tool can supplement science education where access to qualified teachers and individualized support is limited. Future work will involve larger-scale evaluations, deeper partnerships with schools, and the addition of multimodal and tutoring features.

\begin{credits}
\subsubsection{\ackname} We are grateful to ETH for Development (ETH4D) for funding this work with an ETH4D Pilot Grant.
\end{credits}

%% file: refs.bib
@inproceedings{boateng2023b,
  title={Real-World Deployment and Evaluation of Kwame for Science, An AI Teaching Assistant for Science Education in West Africa},
  author={Boateng, George and John, Samuel and Boateng, Samuel and Badu, Philemon and Agyeman-Budu, Patrick and Kumbol, Victor},
  booktitle={International Conference on Artificial Intelligence in Education},
  year={2024},
  organization={Springer}
}

@article{chang2023,
title={Using an artificial intelligence chatbot in scientific inquiry: Focusing on a guided-inquiry activity using inquirybot},
author={Chang, Jina and Park, Joonhyeong and Park, Jisun},
journal={Asia-Pacific Science Education},
volume={9},
number={1},
pages={44--74},
year={2023},
publisher={Brill}
}

@article{chen2024,
title={Effectiveness of AI-assisted game-based learning on science learning outcomes, intrinsic motivation, cognitive load, and learning behavior},
author={Chen, Ching-Huei and Chang, Ching-Ling},
journal={Education and Information Technologies},
volume={29},
number={14},
pages={18621--18642},
year={2024},
publisher={Springer}
}

@article{deveci2021,
title={Chatbot application in a 5th grade science course},
author={Deveci Topal, Arzu and Dilek Eren, Canan and Kolburan Ge{\c{c}}er, Aynur},
journal={Education and Information Technologies},
volume={26},
number={5},
pages={6241--6265},
year={2021},
publisher={Springer}
}

@article{fayzullina2025,
title={Artificial intelligence in science education: A systematic review of applications, impacts, and challenges},
author={Fayzullina, Albinа R and Filippova, Alla A and Garnova, Natalya Y and Astakhov, Dmitry V and Kalmazova, Nadezhda and Zaripova, Zulfiya F},
journal={Contemporary Educational Technology},
volume={17},
number={4},
pages={ep613},
year={2025},
publisher={Bastas}
}

@misc{foondamate,
title = {FoondaMate},
howpublished = {\url{https://foondamate.com/}},
note = {Accessed: 2026-02-03},
year = {2026},
url = {https://foondamate.com/}
}

@inproceedings{henkel2024,
  title={Effective and scalable math support: Experimental evidence on the impact of an AI-math tutor in Ghana},
  author={Henkel, Owen and Horne-Robinson, Hannah and Kozhakhmetova, Nessie and Lee, Amanda},
  booktitle={International conference on artificial intelligence in education},
  pages={373--381},
  year={2024},
  organization={Springer}
}

@article{kurniawan2024,
title={A hybrid automatic scoring system: Artificial intelligence-based evaluation of physics concept comprehension essay test},
author={Kurniawan, W and Riantoni, C and Lestari, N and Ropawandi, D},
journal={International Journal of Information and Education Technology},
volume={14},
number={6},
pages={876--882},
year={2024}
}

@article{lee2023,
title={Improving science conceptual understanding and attitudes in elementary science classes through the development and application of a rule-based AI Chatbot},
author={Lee, Juyeon and An, Taesoo and Chu, Hye-Eun and Hong, Hun-Gi and Martin, Sonya N},
journal={Asia-Pacific Science Education},
volume={9},
number={2},
pages={365--412},
year={2023},
publisher={Brill}
}

@article{lin2023,
author = {Yen-Ting Lin and Jian-Heng Ye},
title = {Development of an Educational Chatbot System for Enhancing Students’ Biology Learning Performance},
journal = {Journal of Internet Technology},
volume = {24},
number = {2},
year = {2023},
issn = {2079-4029}, pages = {275--281}, url = {https://jit.ndhu.edu.tw/article/view/2867}
}

@inproceedings{reimers2019,
  title={Sentence-bert: Sentence embeddings using siamese bert-networks},
  author={Reimers, Nils and Gurevych, Iryna},
  booktitle={Proceedings of the 2019 conference on empirical methods in natural language processing and the 9th international joint conference on natural language processing (EMNLP-IJCNLP)},
  pages={3982--3992},
  year={2019}
}

@misc{rori,
title = {Rori: AI‑Powered Virtual Math Tutor},
howpublished = {\url{https://rori.ai/}},
note = {Accessed: 2026‑02‑03. Rori is an AI‑powered virtual math tutor developed by Rising Academies for improving mathematics learning outcomes.},
year = {2026},
url = {https://rori.ai/}
}

@article{taani2025,
title={ChatGPT in education: Benefits and challenges of ChatGPT for mathematics and science teaching practices},
author={Taani, Osama and Alabidi, Suzan},
journal={International Journal of Mathematical Education in Science and Technology},
volume={56},
number={9},
pages={1748--1777},
year={2025},
publisher={Taylor \& Francis}
}

@misc{UNESCO2021,
title = {The Persistent Teacher Gap in Sub-Saharan Africa is Jeopardizing Education Recovery},
author = {{UNESCO}},
year = {2021},
month = {July},
day = {26},
howpublished = "\href{https://www.unesco.org/en/articles/persistent-teacher-gap-sub-saharan-africa-jeopardizing-education-recovery}{https://www.unesco.org/en/articles/persistent-teacher-gap-sub-saharan-africa-jeopardizing-education-recovery}",
note = {Accessed: 2026-02-02}
}

@misc{dw2020,
  key = "DW",
  title = {Why mobile internet is so expensive in Africa (2020)},
  howpublished = "\href{https://www.dw.com/en/why-mobile-internet-is-so-expensive-in-some-african-nations/a-55483976}{https://www.dw.com/en/why-mobile-internet-is-so-expensive-in-some-african-nations/a-55483976}",
  month =        Nov,
  year =         2020,
  lastaccessed = "Feb, 2024"
}

@misc{UNESCO2024,
title = {Global Report on Teachers: Addressing Teacher Shortages and Transforming the Profession},
author = {{UNESCO} and {International Task Force on Teachers for Education 2030}},
year = {2024},
institution = {UNESCO},
type = {Report},
url = {https://www.teachertaskforce.org/sites/default/files/2024-02/2024_TTF-UNESCO-Global-Report-on-Teachers_EN.pdf},
note = {See p.~50},
}
